\documentclass[sigconf,nonacm]{acmart}

\AtBeginDocument{%
  }

\copyrightyear{2024}
\acmYear{2024}
\acmDOI{XXXXXXX.XXXXXXX}

\usepackage{amsmath,graphicx,amsfonts,amsthm}
\usepackage{algorithmic}
\usepackage{algorithm}
\usepackage{array}
\usepackage{textcomp}
\usepackage{stfloats}
\usepackage{url}
\usepackage{verbatim}
\usepackage{graphicx}
\usepackage{subfigure}
\usepackage{utfsym,colortbl}
\usepackage{tikz}
\usetikzlibrary{arrows, positioning, shapes.geometric}

\definecolor{mygray}{gray}{.92}

\begin{document}

\title{GDGS: Gradient Domain Gaussian Splatting for Sparse Representation of Radiance Fields}

\author{Yuanhao Gong}
\email{gong@szu.edu.cn}
\orcid{0000-0001-5702-1927}
\affiliation{%
  \institution{Electronics and Information Engineering, Shenzhen University, China}
  \country{}
}

\renewcommand{\shortauthors}{Yuanhao et al.}

\begin{abstract}
The 3D Gaussian splatting methods are getting popular. However, they work directly on the signal, leading to a dense representation of the signal. Even with some techniques such as pruning or distillation, the results are still dense. In this paper, we propose to model the gradient of the original signal. The gradients are much sparser than the original signal. Therefore, the gradients use much less Gaussian splats, leading to the more efficient storage and thus higher computational performance during both training and rendering. Thanks to the sparsity, during the view synthesis, only a small mount of pixels are needed, leading to much higher computational performance ($100\sim 1000\times$ faster). And the 2D image can be recovered from the gradients via solving a Poisson equation with linear computation complexity. Several experiments are performed to confirm the sparseness of the gradients and the computation performance of the proposed method. The method can be applied various applications, such as human body modeling and indoor environment modeling. 
\end{abstract}



%  \includegraphics[width=\textwidth]{}
 % \label{fig:teaser}

\maketitle

\section{Introduction}
Getting 3D signals from multi-view images is a key task in the wide-reaching field of computer vision. It involves the intricate job of examining and interpreting the varying viewpoints presented by the array of images, all to build a precise 3D representation of the subject. This core task underpins a host of applications and studies in computer vision, making it a crucial area to grasp and comprehend.

The challenge here is to understand depth and perspective from two-dimensional images, which is not an easy task. It's more than just viewing images - it's about turning flat visuals into a three-dimensional perspective. This involves advanced math techniques that are complex and sophisticated. Strong algorithms are also needed to manage this translation process. They need to interpret and analyze data quickly and accurately. On top of that, we need significant computational power to process the large amount of data and operations. This makes the task complex and challenging, but also quite interesting from a technical perspective.

NeRF, which stands for Neural Radiance Fields, represents a groundbreaking and innovative method in the field of 3D modelling. It employs the use of a fully connected deep network, a complex and intricate system, to model the volumetric scene function. This function is integral to creating a realistic and immersive 3D environment. What sets NeRF apart is its ability to generate high-quality, novel views of 3D scenes. This is achieved from sparse input views, meaning that even with limited input data, the system can produce detailed and comprehensive 3D scene representations. This illustrates the power and potential of the NeRF method in transforming the way we approach and utilize 3D modelling technology.

\begin{figure}
{\includegraphics[width=0.7\linewidth]{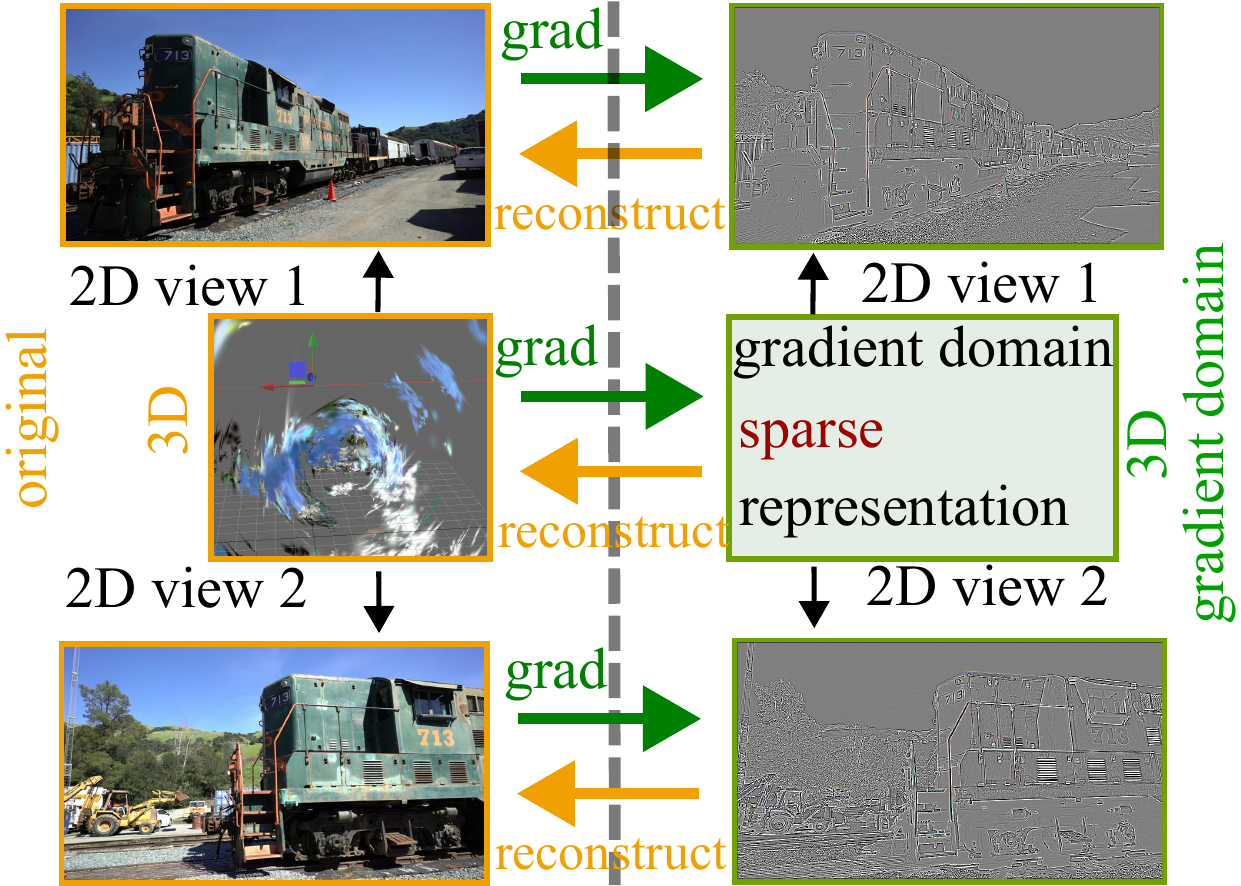}}
	\caption{The relationship between the original signal (left) and the gradient domain representation (right). The gradient domain representation is {\color{red}sparse} and thus only large gradients are needed. As a result, the 2D view gradient is also sparse and only a small part of pixels are rendered. And the final image is reconstructed via solving a Poisson equation.}
	\label{fig:pipe}
\end{figure}

Recently, 3D Gaussian splatting has gained quite a bit of traction~\cite{Kerbl2023}. It's a key player in scene estimation and rendering jobs. It uses the Gaussian distribution to figure out the scene's properties, and then uses this information to create detailed and precise illustrations. It avoids the ray tracing in the NeRF methods. Instead, it uses splatting for the image rendering. This method has been a big step in pushing 3D graphics forward.

\subsection{Particle Representation}
For a 3D signal $f(\vec{x})$, where $\vec{x}$ is the spatial coordinate, it can be expressed as a convolution operation with the classical Dirac delta function
\begin{equation}
	f(\vec{x})=\int f(\vec{\tau})\delta(\vec{x}-\vec{\tau})\mathrm{d}\vec{\tau}\,.
\end{equation} Although this is exact, the abstract delta function is not computationally practical.

To improve the computation property, the above equation is relaxed into the follow discrete expression
\begin{equation}
	\tilde{f}(\vec{x})\equiv\sum_{k=1}^{K}f(\vec{\tau}_k)W(\vec{x}-\vec{\tau}_k, h_k)V_k\,,
\end{equation}where $\tilde{f}$ is the reconstructed signal from this discrete representation, $K$ is the total number of particles, $k$ is the particle index, $W$ is a particle kernel function, $h_k$ is the kernel parameter, and $V_k$ is the volume of the particle.

In most of cases, the multiplication value $f(\vec{\tau}_k)V_k$ can be treated as one variable $A_k$ for convenience reason, leading to the following particle representation
\begin{equation}
	\label{eq:pr}
	\tilde{f}(\vec{x})\equiv\sum_{k=1}^{K}A_kW(\vec{x}-\vec{\tau}_k, h_k)\,.
\end{equation} The introduced $A_k$ can carry multiple features, such as mass, temperature, curvature, etc. It is generic for the particle representation.

With such particle representation, we can evaluate the distance between the original signal $f(\vec{x})$ and its reconstruction from the particle representation. More specifically, the distance is
\begin{equation}
	{L}(f,\tilde{f})=\frac{1}{2}\|f(\vec{x})-\tilde{f}(\vec{x})\|^2_2\,.
\end{equation}
One important property of particle representation is that this distance can be reduced if more particles are added. This property makes the particle representation flexible and compact.

The gradients of the particle representation with respect to the spatial coordinate, the parameter $h_k$ and $\tau_k$ are
\begin{eqnarray}
	\frac{\partial L}{\partial \vec{x}}&=&(\tilde{f}-f)\sum_{k=1}^{K}A_k \frac{\partial W(\vec{x}-\vec{\tau}_k, h_k)}{\partial \vec{x}},\\\
		\frac{\partial L}{\partial h_k}&=&(\tilde{f}-f)\sum_{k=1}^{K}A_k \frac{\partial W(\vec{x}-\vec{\tau}_k, h_k)}{\partial h_k}\,,\\
		\frac{\partial L}{\partial \vec{\tau}}&=&(\tilde{f}-f)\sum_{k=1}^{K}A_k \frac{\partial W(\vec{x}-\vec{\tau}_k, h_k)}{\partial \vec{\tau}}\,.
\end{eqnarray} These gradients can be used to update the center and shape parameters of the particles.
\subsection{3D Gaussian Splatting}
The 3D Gaussian splatting (3DGS) method and its variants are special cases of the particle representation. More specifically, the 3D Gaussian splatting uses the anisotropic Gaussian kernels in the particle representation
\begin{eqnarray}
	\label{eq:3dgs}
	\tilde{f}(\vec{x})&=&\sum_{k=1}^{K}A_kG(\vec{\tau}_k,\Sigma_k)\,,\\
	\mathrm{where}\, G(\vec{\tau}_k,\Sigma_k)&=&\exp[-(\vec{x}-\vec{\tau}_k)^T\Sigma_k^{-1}(\vec{x}-\vec{\tau}_k)]\,.
\end{eqnarray}The non negative covariance matrix is $\Sigma=RSS^TR^T$, where $S$ is a diagonal scaling matrix and $R$ is a rotation matrix.

This 3D Gaussian particle is then splatted on the the 2D image plane. The covariance in 2D is computed via
\begin{equation}
	\label{eq:sig2d}
	\Sigma^{2D}=JW\Sigma W^TJ^T\,,
\end{equation} where $W$ is the world-to-camera matrix and $J$ is a local matrix for the projection.

The color $c(u,v)$ at a view is then defined via an alpha blending
\begin{equation}
	\label{eq:color}
	c(u,v)=\sum_{k=1}^{K}c_k\alpha_kG^{2D}_k\prod_{j=1}^{k-1}(1-\alpha_jG_j^{2D})\,,
\end{equation}where $c_k$ is a view dependent color, $\alpha_k$ is the transparency, $G^{2D}$ is a 2D Gaussian function with the covariance matrix $\Sigma^{2D}$ in Eq.~\eqref{eq:sig2d}. And $k$ is sorted from the view direction. The coordinate $(u,v)$ indicates the image space coordinate.

The rendered image $c(u,v)$ is then compared with the observed image $im$ via a $\ell_1$ distance and the structural similarity distance
\begin{equation}
	\label{eq:origloss}
	Loss(c,im)=(1-\lambda)|c-im|+\lambda D_{SSIM}(c,im)\,,
\end{equation} where $\lambda>0$ is a weight parameter and $D_{SSIM}$ is a distance measurement using $SSIM$.

\subsection{Variants of Gaussian Splatting}
Thanks to the splatting, the Gaussian splatting methods do not need the ray tracing to render the observed image at a given view. As a result, these methods are much faster than the Nerf based approaches that require the ray tracing to perform the image rendering.

With the advantages of computational efficiency and the resulting high quality rendered images, various Gaussian splatting methods have been developed. For example, it can be applied in street modeling~\cite{yan2023streetgaussians}, human head modeling~\cite{wang2024gaussianhead,zhou2024headstudio} and human body modeling~\cite{abdal2023gaussian}. In~\cite{gong2024isotropic}, isotropic Gaussian function is adopted to reduce the orientation issue. And A compression method is developed to remove the unimportant Gaussian particles, reducing the file size~\cite{lee2023compact,fan2024lightgaussian}. When we are writing this paper, 2D Gaussian splatting method is developed in~\cite{huang20242d}, where 2D disks are attached to a surface. Similar idea is also shown in~\cite{guedon2023sugar}. This method is more suitable for surface representation instead of volume representation. In~\cite{gong2024eggs}, an edge guided Gaussian splatting method is developed, forcing the particle to be aligned with edges in the image. A survey on 3D Gaussian splatting can be found in~\cite{chen2024survey}.

\subsection{Gradient Domain Signal Processing}
Gradient domain image processing is a technique that works with the image's gradient field~\cite{gong:gdp,gong2023gradient}, rather than the image itself. The relationship between the original and gradient domain signals is shown in Fig.~\ref{fig:pipe}. The gradients have been hugely useful in various image processing tasks like seamless image stitching, image in-painting, and tone mapping. By using the gradient domain, we can keep the structural details of images, even as we make significant changes that might be lost if we worked directly on the image.

This process has two steps: firstly, we apply the desired changes to the image's gradient, creating a new gradient field. Then, we recreate an image that matches this new gradient field as closely as possible. This second step is all about optimization and often uses methods like Poisson reconstruction.

Moreover, gradient domain processing also enables the preservation of fine details in an image or signal while reducing noise, making it ideal for applications like image editing, video processing, and computer vision.

Furthermore, this processing technique is remarkably versatile. It can be applied in a vast array of fields ranging from medical imaging to machine learning, opening up new possibilities for innovation. For instance, in medical imaging, gradient domain processing could help enhance the visibility of certain structures or abnormalities, thereby assisting in more accurate diagnostics. 

Mesh processing, meanwhile, is to manipulate a polygonal or polyhedral mesh. This is a collection of vertices, edges, and faces that make up a 3D object's shape. It's a crucial process in areas like computer graphics, computer-aided design, and virtual reality.

There are many types of mesh processing, such as mesh smoothing (to remove noise), mesh simplification (for quicker rendering), and mesh parameterization (to map a texture onto a mesh). Much like gradient domain image processing, mesh processing techniques often aim to optimize a particular property of the mesh, such as its smoothness, compactness, or the distortion of the texture mapping.

To sum up, both gradient domain image processing and mesh processing are potent tools in their respective areas, enabling complex changes while preserving key details. These methods are at the heart of image and graphic processing, driving progress in fields ranging from digital photography to video game development.

\begin{figure}
	\subfigure[one image]{\includegraphics[height=0.2\linewidth]{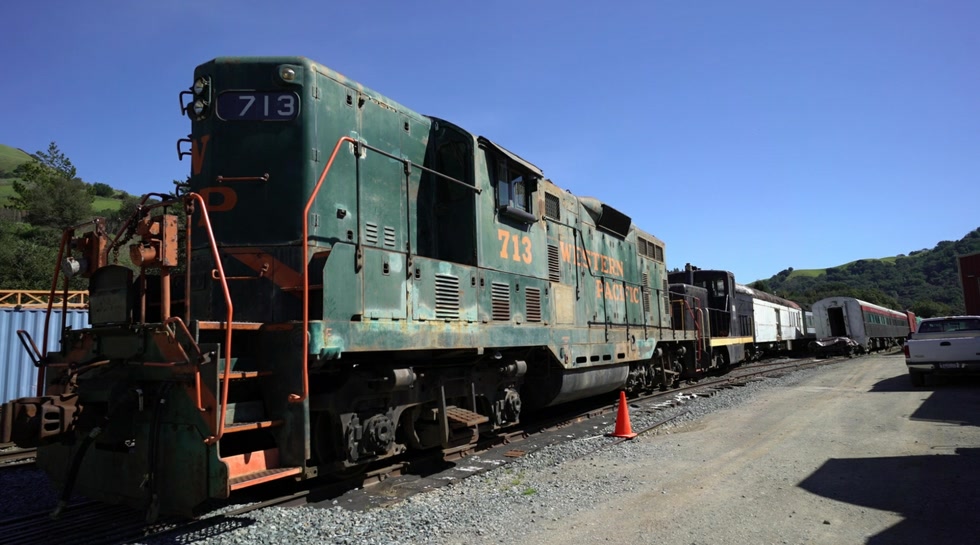}}
	\subfigure[pdf of the image]{\includegraphics[width=0.27\linewidth]{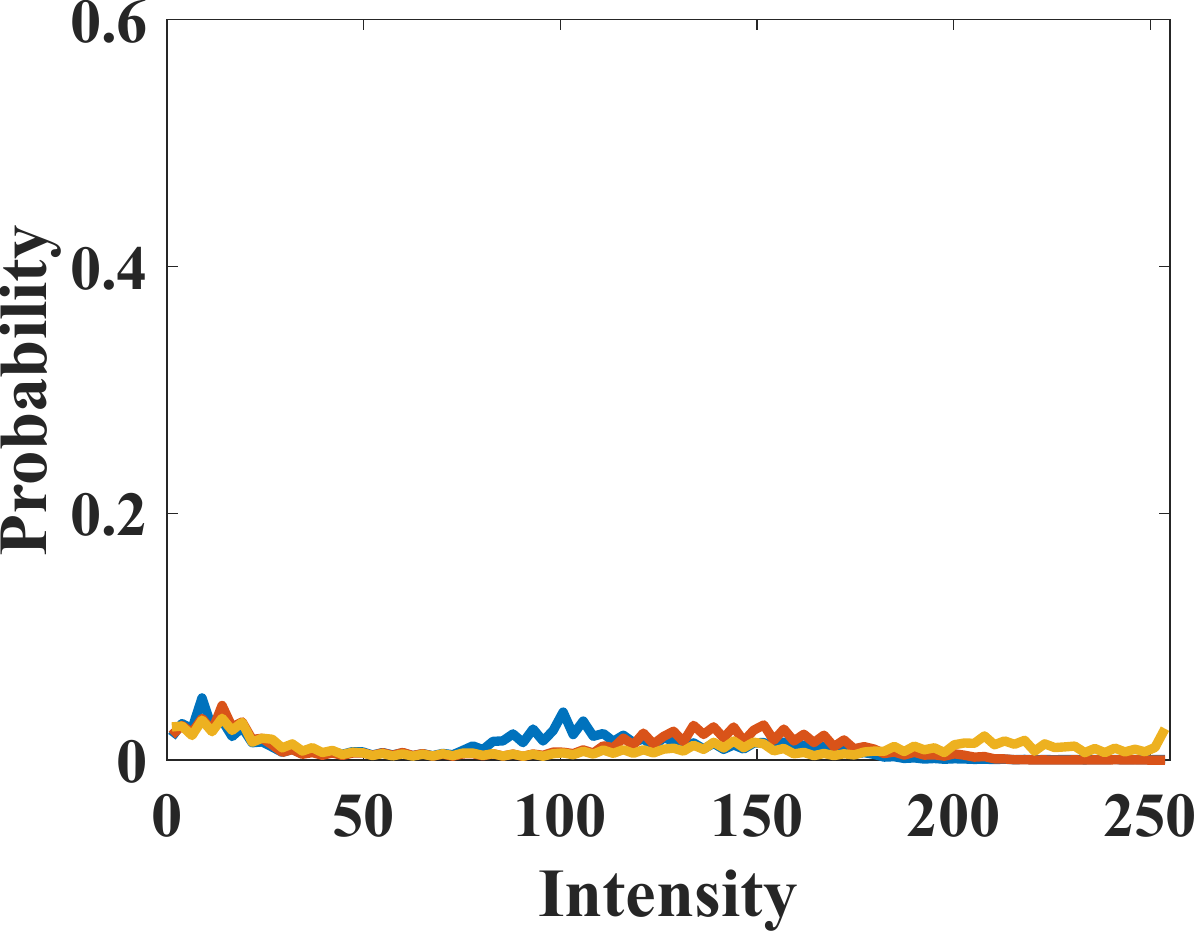}}
	\subfigure[cdf of the image]{\includegraphics[width=0.27\linewidth]{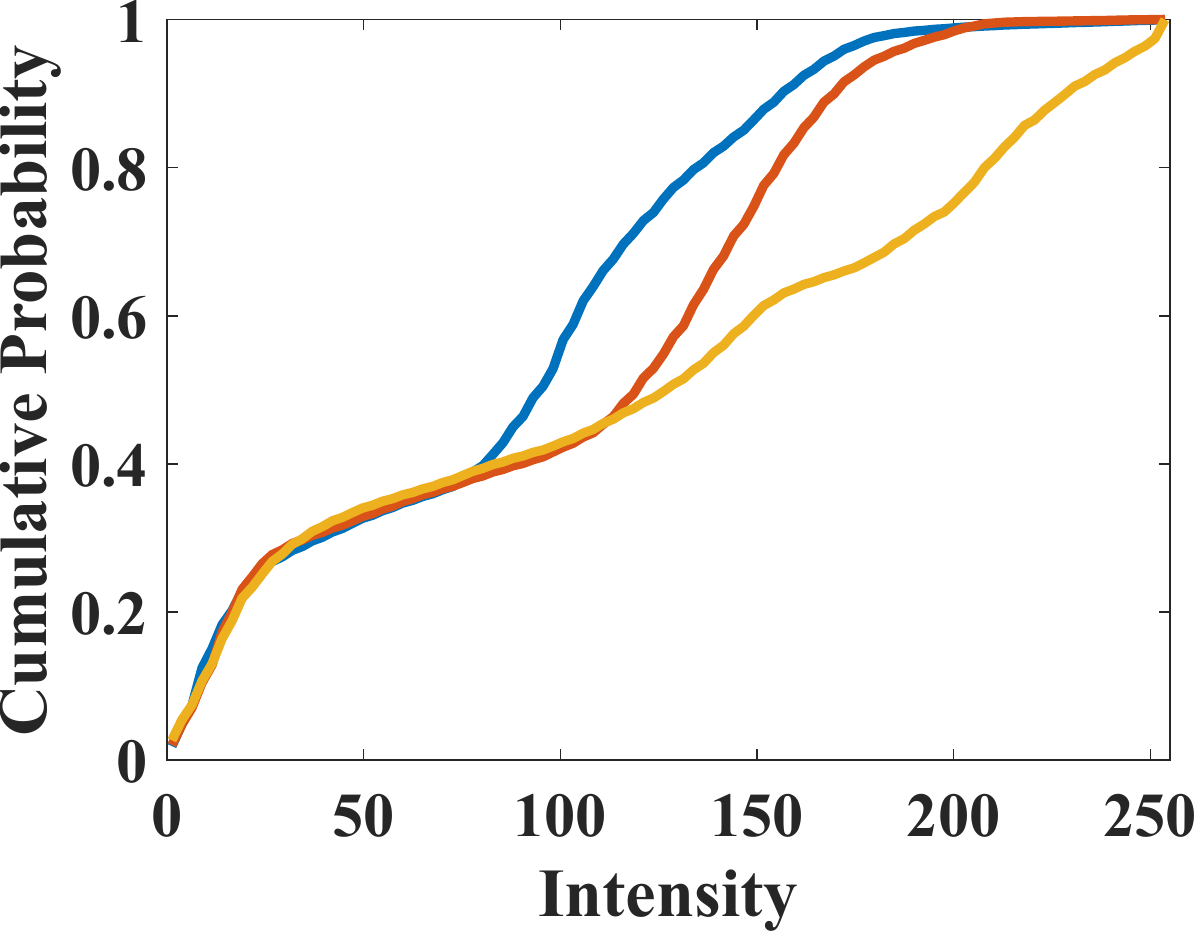}}
	
	\subfigure[Laplacian field (scaled and shifted for better visualization)]{\includegraphics[height=0.2\linewidth]{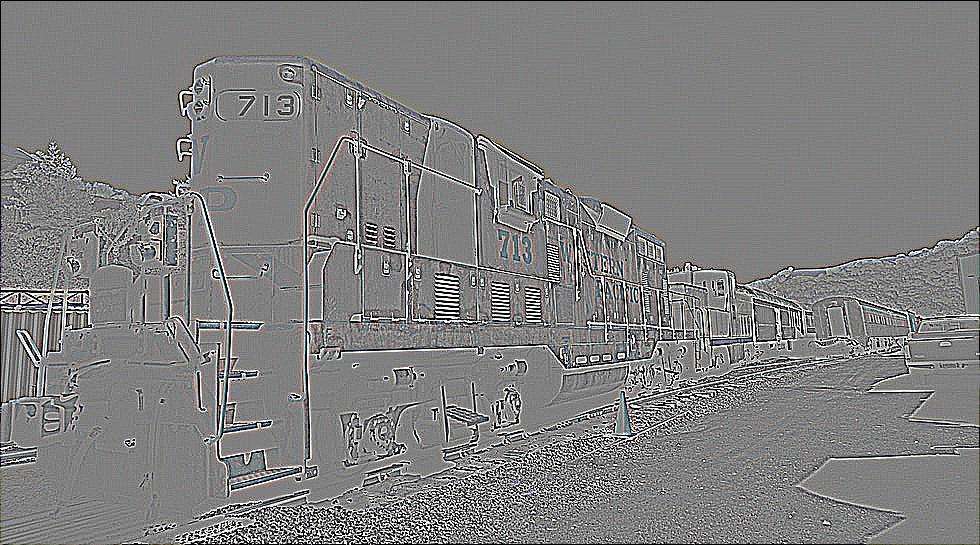}}
	\subfigure[pdf of the Lap]{\includegraphics[width=0.27\linewidth]{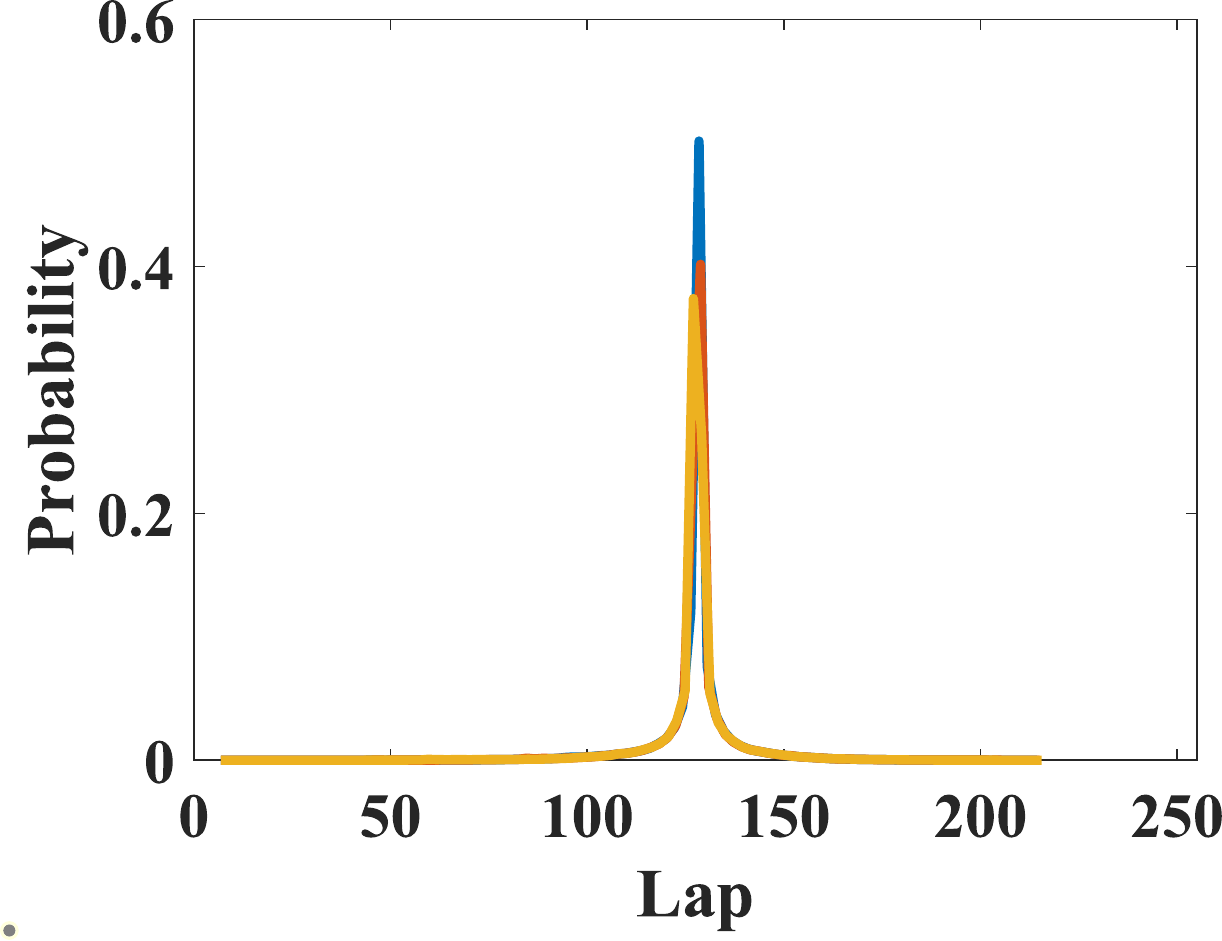}}
	\subfigure[cdf of the Lap]{\includegraphics[width=0.27\linewidth]{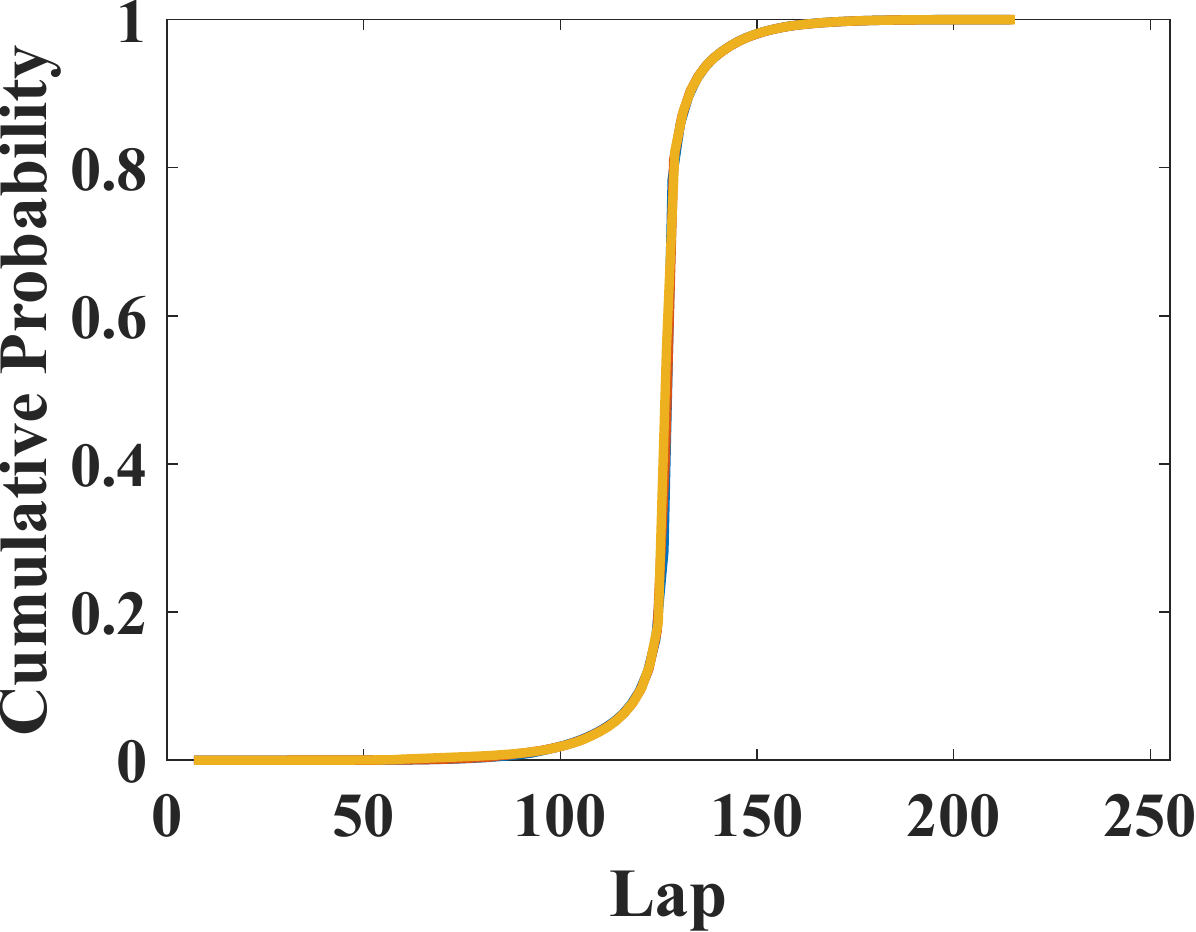}}
	\caption{A 2D image (a) and its Laplace field (d). The middle and right columns show their probabilities and cumulative probabilities, respectively, where different color indicate color channels. The values are shifted with 128 but without scaling. As these figures shown, the Laplacian field is much sparser than the original image. Therefore, in the Laplacian field, we do not necessarily render all the pixels as in the 3D Gaussian splatting. Instead, only sparse edges are computed. Meanwhile, the rendered image can be recovered by its Laplacian field via solving a Poisson equation.}
	\label{fig:1}
\end{figure}
\subsection{Sparsity}
A key advantage of gradient domain signal processing is its sparsity~\cite{gong2023gradient}. This characteristic allows for a more efficient signal representation, cutting down on the data that needs processing. This boosts computational speed and saves on storage space when one is working with limited processing power or storage. 

Although there are three components in the gradients for 3D signals, we will show that only the Laplacian field (the divergence of the gradients) matters. The Laplacian field is even sparser than the gradient, because the constant gradients become zero after the divergence operator.

One example is shown in Fig.~\ref{fig:1}, where the Laplacian field (divergence of the gradient field) is much sparser than the original signal. The sparsity becomes even higher in the 3D space~\cite{gong:gdp,gong:phd}.
\subsection{Our Motivation and Contributions}
Different from previous Gaussian splatting methods that directly work on the original signal, we propose to model the gradients of the signal with Gaussian splats. As mentioned, gradient domain has several advantages. One of them is the sparsity. Therefore, the resulting gradient domain Gaussian splats are much sparser than the original Gaussian splats. Our contributions are
\begin{itemize}
	\item  we introduce the gradient domain Gaussian splatting method.
	\item  the gradient domain Gaussian splatting is much sparser than the original one.
	\item the synthetic view 2D image can be exactly recovered by its gradients via solving a Poisson equation.
\end{itemize}
\section{Gradient Domain Gaussian Splatting}
In this paper, instead of approximating the original signal $f(\vec{x})$, we study its gradient $\nabla f(\vec{x})$, where $\nabla$ is the gradient operator. In most of cases, the measurement error satisfies a Gaussian distribution and thus we can use the $\ell_2$ norm to measure the reconstruction error in the gradient domain
\begin{equation}
	{L_g}(f,\tilde{f})=\frac{1}{2}\|\nabla f(\vec{x})-\nabla \tilde{f}(\vec{x})\|^2_2\,.
\end{equation} Minimizing this equation leads to the following Poisson equation
\begin{equation}
	\Delta f(\vec{x})=\Delta \tilde{f}(\vec{x})\,,
\end{equation} where $\Delta=\nabla\cdot\nabla=\frac{\partial^2}{\partial x^2}+\frac{\partial^2}{\partial y^2}+\frac{\partial^2}{\partial z^2}$ is the Laplacian operator.

Above two equations link the gradient domain processing with the well-known Poisson equation, which has preferred theoretical mathematical properties, such as the unique solution and efficient numerical solvers. Moreover, the Laplacian field is sparser than the gradient field, leading to higher computational performance.

There are many efficient numerical solvers for the Poisson equation and they can achieve linear computational complexity. Roughly speaking, they can be considered as direct solvers or iterative solvers. Direct solvers, like the name implies, typically use factorization methods to solve a smaller, denser system matrix, such as Gaussian elimination and LU decomposition. Iterative solvers deal with the large and sparse system matrix. They start with an initial estimate and gradually refine it to closer to the solution, such as Jacobi Method, Gauss-Seidel Method and Multigrid Methods.
\subsection{Poisson Equation}
Let us define the Laplacian field of the original signal as
\begin{equation}
	\rho(\vec{x})\equiv \Delta f(\vec{x})\,.
\end{equation} If $\rho(\vec{x})$ is given, the signal $f(\vec{x})$ can be reconstructed via solving the following Poisson equation with a proper boundary condition (Neumann boundary condition in most cases)
\begin{equation}
	\Delta u(\vec{x})=\rho(\vec{x})\,
\end{equation} where $u(\vec{x})$ is the unknown signal to be recovered. Mathematically, $u(\vec{x})=f(\vec{x})$ (with possible constant difference). In practice, although there might be some numerical error, the error is not obviously visible, especially for vision tasks. For example, when $\|u-f\|_{\infty}=0.01$, there is no visual difference in practical applications. And $u$ is an accurate reconstruction of $f$.

If $\rho(\vec{x})$ is obtained from the real physical world, then $u(\vec{x})$ must exist. From mathematical theories, we know that the solution from the Poisson equation is unique. Such uniqueness gives theoretical guarantee about this equation.

Be aware that we do not necessarily solve the Poisson equation in 3D to get the signal $f(\vec{x})$. Instead, we are focusing on the image synthesis. Therefore, we will solve the 2D Poisson equation after the view projection step.
\subsection{Our Method}
Our gradient domain Gaussian splatting method has three steps. First, we approximate the Laplacian field of the original signal with the Gaussian splats. Then, we project these splats onto the image plane, obtaining the 2D Laplacian field. Finally, an image is reconstructed from the 2D Laplacian field by solving a Poisson equation. The complete algorithm is summarized in Algorithm~\ref{algo} and the details are explained in the following subsections.
\subsubsection{3D Laplacian Field}
Instead of approximating the original signal $f(\vec{x})$, we propose to approximate its Laplacian field $\rho(\vec{x})\equiv\Delta f(\vec{x})$ via the Gaussian particles. More specifically, we have
\begin{eqnarray}
	\label{eq:gdgs}
	\tilde{\rho}(\vec{x})&=&\sum_{k=1}^{K}A_kG(\vec{\tau}_k,\Sigma_k)\,,\\
	\mathrm{where}\, G(\vec{\tau}_k,\Sigma_k)&=&\exp[-(\vec{x}-\vec{\tau}_k)^T\Sigma_k^{-1}(\vec{x}-\vec{\tau}_k)]\,.
\end{eqnarray}
\subsubsection{Projection}
Then, we project these 3D Gaussian kernels into the 2D image plane with the following 2D covariance matrix
\begin{equation}
	\Sigma^{2D}=JW\Sigma W^TJ^T\,,
\end{equation} where $W$ is the world-to-camera matrix and $J$ is a local affine matrix for the projection. And the Laplacian field of a 2D image in a view direction can be obtained via the alpha composition
\begin{equation}
	\label{eq:gdgs2d}
	\tilde{\rho}^{2D}(u,v)=\sum_{k=1}^{K}c_k\alpha_kG^{2D}_k\prod_{j=1}^{k-1}(1-\alpha_jG_j^{2D})\,.
\end{equation}
Be aware that $\tilde{\rho}^{2D}(u,v)$ is sparse and only a small number of pixels are needed.
\subsubsection{Poisson Equation}
Finally, the color image can be recovered via solving the following Poisson equation with Neumann boundary condition
\begin{equation}
	\label{eq:p2d}
	\Delta c(u,v)=\tilde{\rho}^{2D}(u,v)\,.
\end{equation}

This equation is a 2D Poisson equation for an image reconstruction. It can be efficiently solved by methods with linear computation complexity, such as multi-grid method and convolution pyramids~\cite{Farbman2011,gong:phd}. It can also solved by convolution neural networks (CNN)~\cite{Aggarwal2019,li2021fourier, Oezbay2021,Peng2023}. Choosing traditional methods or CNN depends on several things such as the image resolution, computation resource and required running time. We will discuss the reconstruction in later section.
\begin{algorithm}[!tb]
	\caption{Gradient Domain Gaussian Splatting}
	\label{algo}
	\begin{algorithmic}
		\REQUIRE input images $im$, camera poses
		\WHILE{not converge}
		\STATE update the Laplacian field $\tilde{\rho}(\vec{x})$ via Eq.~\eqref{eq:gdgs}
		\STATE update projected field $\tilde{\rho}^{2D}$ via Eq.~\eqref{eq:gdgs2d}
		\STATE get the rendered image via solving Eq.~\eqref{eq:p2d}
		\STATE evaluate the Loss in Eq.~\eqref{eq:ourloss}
		\ENDWHILE
		\ENSURE $\rho(\vec{x})$
	\end{algorithmic}
\end{algorithm} 
\subsection{Loss Function}
Sine we work in the gradient domain, we prefer the loss function that has a gradient term for better edge alignment. More specifically, we use the following loss function
\begin{equation}
	\label{eq:ourloss}
	L(c,im)=(1-\lambda)\|(c-im)\|_1+\beta\|\nabla c-\nabla im\|_1+\lambda D_{SSIM}(c,im)\,.
\end{equation} 

The proposed loss function forces the rendered image to be aligned with the observed image in the gradient domain (edges). 
\section{Sparsity and Reconstruction}
One important advantage of the Laplacian fields in 3D and 2D is the sparsity. Such sparsity leads to a higher computation performance in practice. Meanwhile, the sparse Laplacian field can accurately recover the original signal via solving a Poisson equation, for which Poisson solvers are numerically efficient and accurate.

If we use less samples (sparser), then the reconstruction error might be increased (error becomes larger). Such trade-off depends on the complexity of the content, computation complexity, required running time, etc. In most of cases, we use the reconstruction error to automatically control the sparsity.
\subsection{Sparsity}
In the 3D space, comparing the Eq.~\eqref{eq:3dgs} and Eq.~\eqref{eq:gdgs}, we notice that the Eq.~\eqref{eq:gdgs} is sparser. After projection onto the imaging plane, the 2D Laplacian field in Eq.~\eqref{eq:gdgs2d} is also sparser than the original Eq.~\eqref{eq:color}.

To measure the sparsity, we use a Cauchy distribution, whose cumulative probability has the form
\begin{equation}
	Cauchy(x,\gamma)=\frac{1}{\pi}\arctan (\frac{x-x_0}{\gamma})+\frac{1}{2}\,
\end{equation} where $\gamma$ is a scale parameter. The smaller $\gamma$, the sparser the distribution. The parameter $\gamma$ is a rough measurement of the sparsity.

\begin{figure}
	\subfigure[intensity domain]{\includegraphics[width=0.45\linewidth]{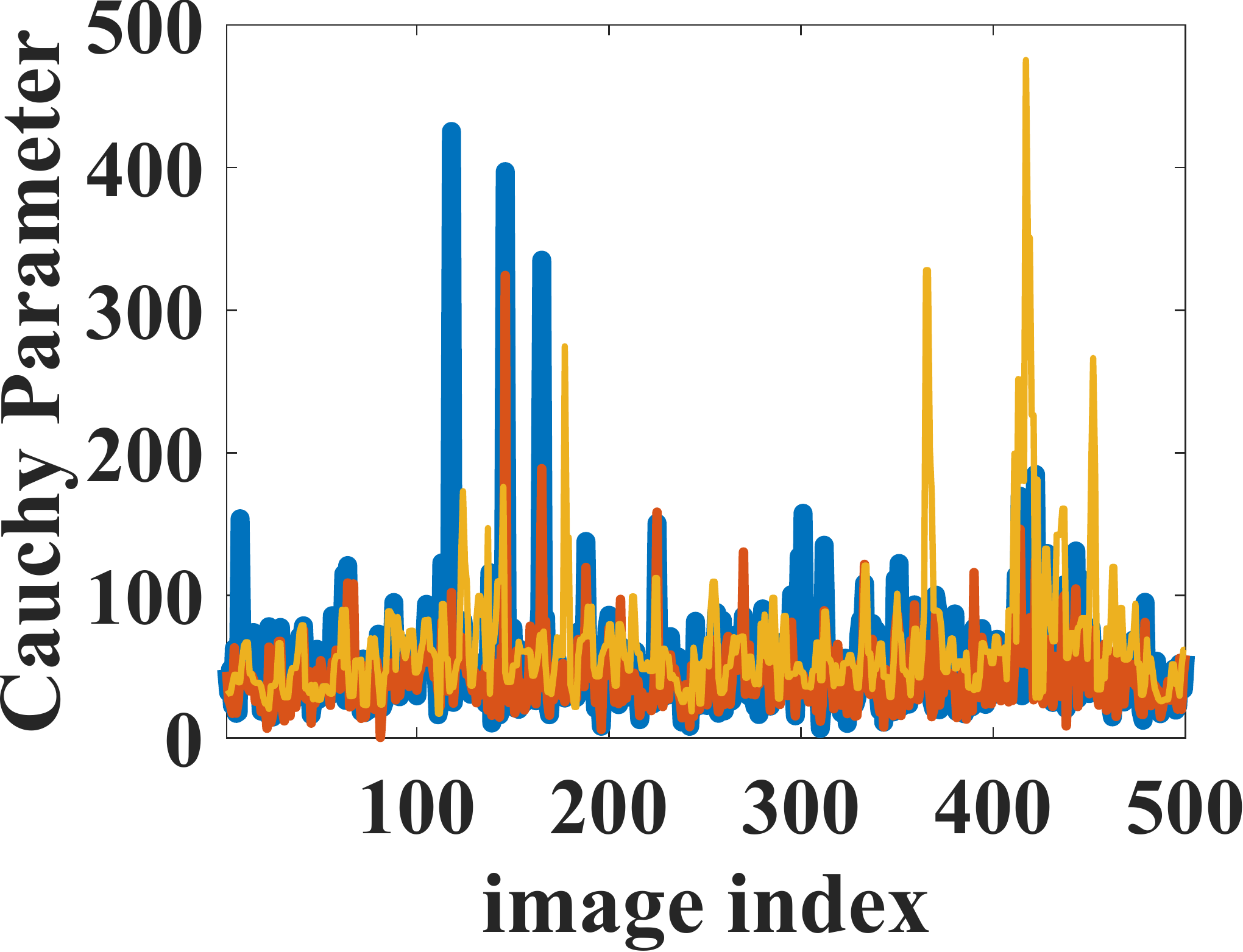}}
	\subfigure[Laplacian domain]{\includegraphics[width=0.45\linewidth]{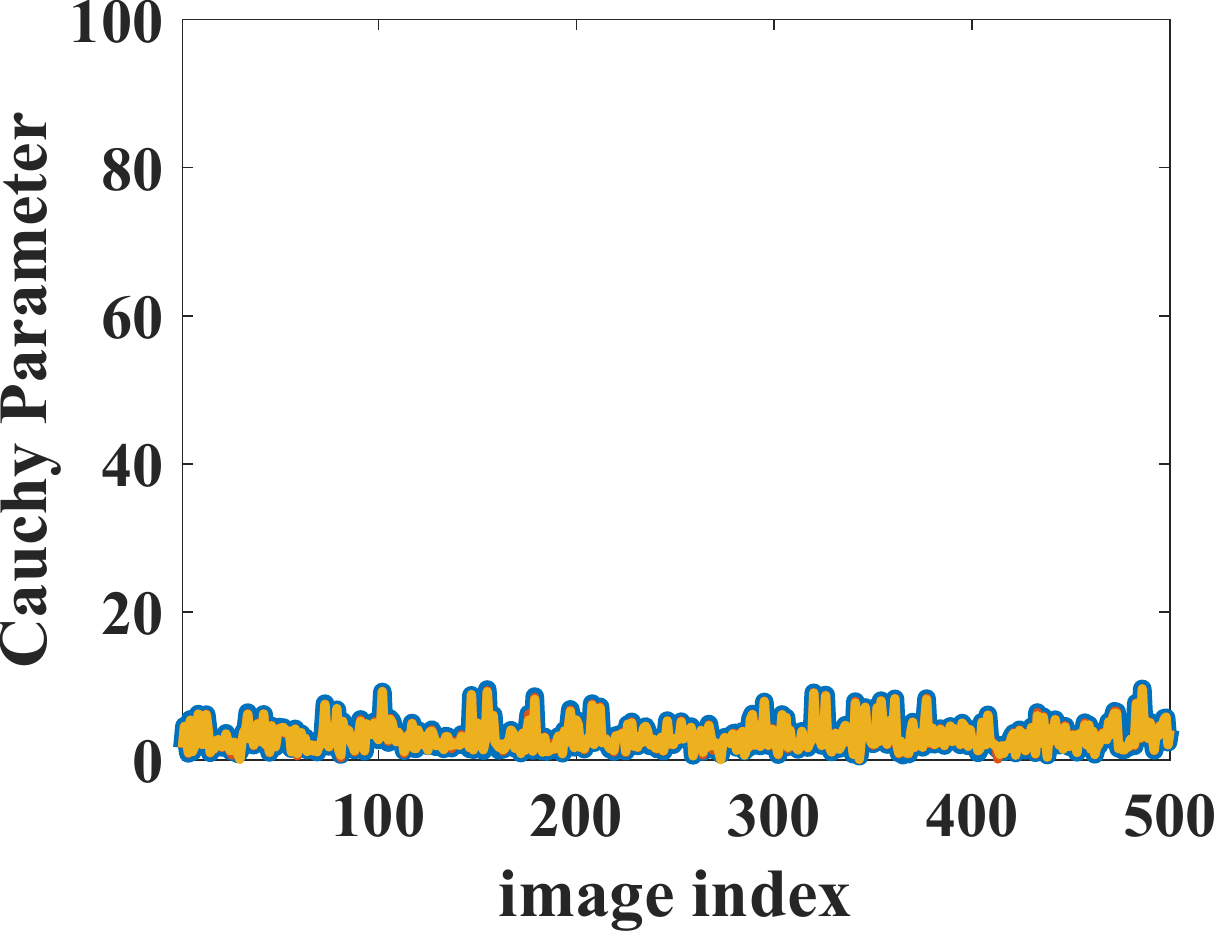}}
	\caption{The Cauchy parameter $\gamma$ in the intensity domain (a) and Laplacian domain (b), respectively. The smaller $\gamma$, the sparser the signal. The mean and median for (a) is 84.4 and 44.6. The mean and median for (b) is 3.1 and 2.8. The Laplacian field is much sparser. The different colors indicate the color channels in RGB images.}
	\label{fig:sparse}
\end{figure}

We model the image intensity and the Laplacian field of 500 images from BSDS500 dataset. And the resulting Cauchy parameter $\gamma$ is shown in Fig.~\ref{fig:sparse}. The mean and median in the intensity domain is 84.4 and 44.6. The mean and median in the Laplacian domain is 3.1 and 2.8. Clearly, the Laplacian field is much sparser. We usually use a threshold function $T$ to obtain the sparse representation
\begin{equation}
	\label{eq:thresh}
	T(x)=\left\{
	\begin{aligned}
		x,&\quad |x|>=t\\
		0,&\quad else\\
	\end{aligned}\right.\,,
\end{equation} where $t$ is a scalar parameter. A larger $t$ leads to a sparser signal representation.

Thanks to the sparsity, the number of Gaussian kernels in our method is significantly smaller than the number of Gaussians in the original signal domain. Such improvement can simplify the representation and accelerate the rendering process, which are important for practical applications.

\subsection{Reconstruction}
After obtaining the sparse $\tilde{\rho}^{2D}$, we reconstruct the corresponding image via solving the Poisson equation, Eq.~\eqref{eq:p2d}. Be aware that this equation is in the 2D image domain, not in the 3D object space.

\subsubsection{Classical Solvers}
The Poisson equation is a classic and essential partial differential equation that is widely used in the fields of mathematical physics and engineering. Solving this equation traditionally involves the use of classical solvers. These solvers primarily employ numerical methods such as finite difference methods, finite element methods, and boundary element methods.

In finite difference methods, the equation is approximated by replacing the derivatives by differences. Finite element methods break down the problem into smaller, simpler parts that are called finite elements. These finite elements are then assembled into a larger system that models the entire problem. The boundary element method, on the other hand, reduces the problem to a boundary only problem, thus significantly reducing the complexity of the problem. All these methods discretize the continuous problem into a system of algebraic equations that can be solved either iteratively or directly.

\begin{figure}
	\subfigure[original]{\includegraphics[width=0.32\linewidth]{image/input1.jpg}}
	\subfigure[t=0.003]{\includegraphics[width=0.32\linewidth]{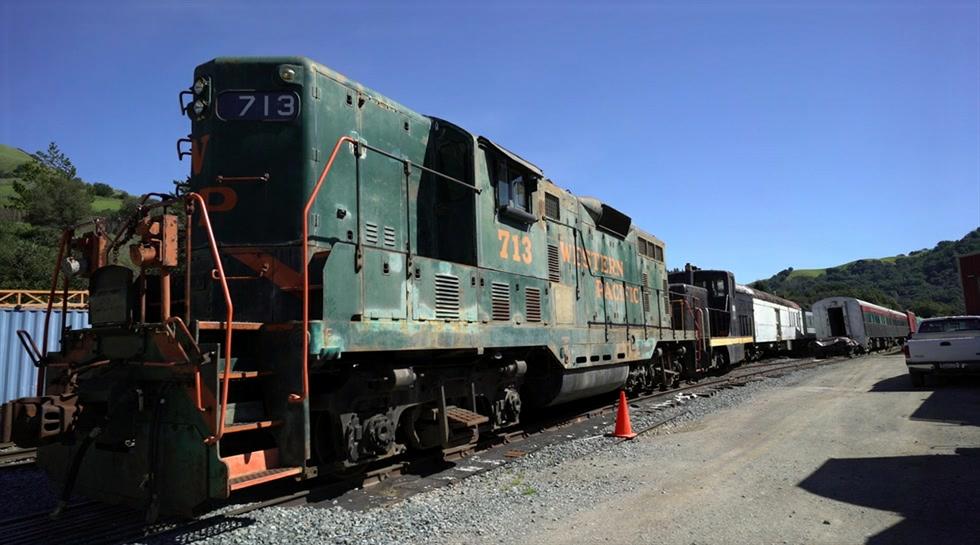}}
	\subfigure[t=0.004]{\includegraphics[width=0.32\linewidth]{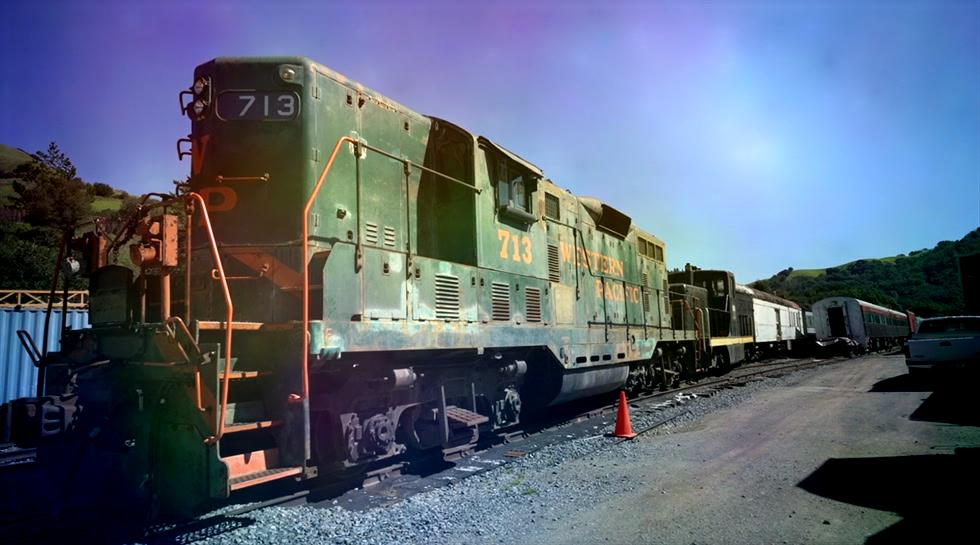}}
	\caption{The traditional Poisson solvers can accurately reconstruct the image as shown in (b). But they might generate artifacts when we force the more sparsity, as shown in (c).}
	\label{fig:sparse2}
\end{figure}

These traditional solvers, however, require the input $\rho$ must be accurate. In other words, these solvers are not robust with the $\rho$. In the scenario of Gaussian splatting, Eq.~\eqref{eq:gdgs2d}, $\rho$ might contain numerical errors, leading to possible failure cases. reconstruction. For example, we can use Eq.~\eqref{eq:thresh} to force the sparsity. When we set $t=0.003$ in Eq.~\eqref{eq:thresh}, $84\%$ of pixels in the Laplacian field $\rho$ are nonzero and the reconstruction MSE is $6\times 10^{-5}$. When we set $t=0.003$ in Eq.~\eqref{eq:thresh}, $62\%$ of pixels in the Laplacian field $\rho$ are nonzero and the reconstruction MSE is $0.013$. These results are visually shown in Fig.~\ref{fig:sparse2}. 
\subsubsection{Deep Learning Solvers}
In contrast to the traditional methods, deep learning methods for solving the Poisson equation take a modern approach. Deep learning leverages the power of neural networks to approximate the solution to the equation. Neural networks are trained using a loss function that minimizes the difference between the predicted solution and the actual solution.

The deep learning approach simplifies the problem by converting it into an optimization problem, thus making it more manageable. Furthermore, this method has shown promise in providing accurate and efficient solutions. One of its greatest advantages is its ability to handle complex geometries and boundary conditions, which are often challenging for classical methods. Therefore, deep learning methods for solving the Poisson equation present an exciting new avenue for solving partial differential equations.

\subsubsection{Our Solver}
\begin{figure}
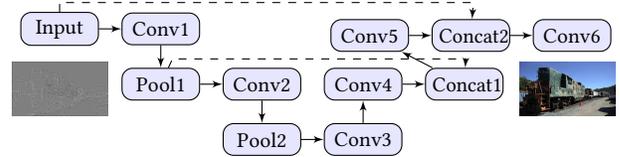

	\begin{tikzpicture}[node distance=0.3cm,scale=0.5]
		\small
		% Define styles for layers
		\tikzset{
			block/.style={rectangle, draw, fill=blue!10, text width=3em, text centered, rounded corners, minimum height=1.5em},
			line/.style={draw, -latex'},
		}
		
		% Draw the layers
		\node[block] (input) {Input};
		\node[block, right=of input] (conv1) {Conv1};
		\node[block, below=of conv1] (pool1) {Pool1};
		\node[block, right=of pool1] (conv2) {Conv2};
		\node[block, below=of conv2] (pool2) {Pool2};
		\node[block, right=of pool2] (conv3) {Conv3};
		\node[block, above=of conv3] (unconv1) {Conv4};
		\node[block, right=of unconv1] (concat) {Concat1};
		\node[block, above left=of concat] (output) {Conv5};
			\node[block, right=of output] (output2) {Concat2};
		\node[block, right=of output2] (output3) {Conv6};
		
		% Draw the lines
		\draw[line] (input) -- (conv1);
		\draw[line] (conv1) -- (pool1);
		\draw[line] (pool1) -- (conv2);
		\draw[line] (conv2) -- (pool2);
		\draw[line] (pool2) -- (conv3);
		\draw[line] (conv3) -- (unconv1);
		\draw[line] (unconv1) -- (concat);
		\draw[line] (concat) -- (output);
		\draw[line] (output) -- (output2);
		\draw[line] (output2) -- (output3);
		
		% Add skip connections
		\draw[line, dashed] (pool1) -- (3,-0.8)--(10.8,-0.8)-- (concat);
		\draw[line, dashed] (input) -- (0,0.7)-- (11,0.7)-- (output2);
		\node at (0,-1.6) {\includegraphics[width=0.15\linewidth]{image/inputL.jpg}};
		\node at (13.5,-1.6) {\includegraphics[width=0.15\linewidth]{image/input1.jpg}};
	\end{tikzpicture}
	\caption{We use a U-Net structure to reconstruct the image from its Laplacian field. We add noise into the input to improve the robustness. We also use the mask in the input to improve the sparsity.}
	\label{fig:unet}
\end{figure}

Following the neural network solvers, we use the U-net structure to solve the Poisson equation. Our network structure is shown in Fig.~\ref{fig:unet}, where the depth depends on the image resolution and the channel number depends on the complexity of the scene. Be aware that this network is a convolution network. Thus, it can be trained on image patches without considering the image resolution restriction.

In general, there are two ways to use this solver. One way is to train a network on millions of images, so the resulting network is generic for all possible scenes. Such trained solver can be adopted for different scenes. Another way is to train the network for each scene separately. Such trained solver is more adaptive to the input images and improves the reconstruction accuracy. It, however, can not be generalized to other scenes. The choice depends on the application. In this paper, we use the second way to get the accuracy.

To improve the performance of our solver, we add noise and sparsity in the input Laplacian field. When we add the noise in the Laplacian field, the robustness is improved. When we use the mask in the input, the network can perform the reconstruction from a sparse input.
	
\subsubsection{A Novel Loss}
For deep learning solvers, the loss function usually is 
\begin{equation}
	L_p=\frac{1}{2}\|\Delta c-\tilde{\rho}^{2D}\|^2\,,
\end{equation} where $c$ is the reconstruction image (the output from the deep learning solvers), and $\tilde{\rho}^{2D}$ is the input sparse signal. This loss function is in the Laplacian domain.

Another loss function is derived from the Euler-Lagrange Variation point of view
\begin{equation}
	L_e=\int_{u,v} [\frac{1}{2}(\nabla c(u,v))^2+c(u,v)\tilde{\rho}^{2D}(u,v)]\mathrm{d}u\mathrm{d}v\,,
\end{equation} whose Euler-Lagrange equation exactly is the target Poisson equation, Eq.~\eqref{eq:p2d}. This loss function works in the original signal domain.

In this paper, we use a hybrid loss with parameter $0\le\theta\le 1$
\begin{equation}
	L_h=\theta L_p+(1-\theta)L_e\,.
\end{equation} This hybrid loss function uses both the Laplacian domain distance and the original signal domain distance. It tries to find the unique solution in both domains. Thus, it is more effective in practice.

\section{Experiments}
In this section, we perform several experiments to confirm that the proposed gradient domain Gaussian splatting method can represent the scene and achieve high quality rendering images. Meanwhile, thanks to the gradient domain, it is much sparser than the original 3D Gaussian splatting method, leading to a more effective representation. The resulting particles are aligned with the edges in the scene, indicating the change of the signal instead of the signal itself.

\subsection{Banana Dataset}
The banana data set contains 16 images at different views. And each image has the $3008\times2000$ resolution. Such high resolution can capture the details in the scene and improve the quality of the radiance field. We use 3DGS and GDGS to perform the reconstruction. The time steps are set as 10K.

We compared the 3DGS  and GDGS on this data set. The largest PSNR for 3DGS is about 41.7dB. In contrast, GDGS can achieve 42.8dB, which is about 1.1dB improvement. Such improvement is a big step to improve the radiance field.

Moreover, thanks to the gradient domain representation, the proposed method is much sparser. After the training, the 3DGS method contains about $357\times 10^3$ particles while the GDGS contains only $3\times 10^3$ particles, which is about 100 times smaller. Such effectiveness also leads to a higher frame rate in the rendering process. 

\subsection{Train Dataset and Truck Dataset}
The train data set contains 301 images, which have the $980\times545$ resolution. The best PSNR for 3DGS and GDGS on this data set is 28.0 and 28.6, respectively. The improvement is about 0.6dB. 

Moreover, thanks to the gradient domain representation, the proposed method is much sparser. After the training, the 3DGS method contains about $372\times 10^3$ particles while the GDGS contains only $2.9\times 10^3$ particles, which is more than 100 times smaller. 

The truck data set contains 251 images, which have the $979\times546$ resolution. It shows similar behavior as previous data sets. The best PSNR for 3DGS and GDGS on this data set is 28.4 and 29.3, respectively. The improvement is about 0.9dB. 

After the training, the 3DGS method contains about $558\times 10^3$ particles while the GDGS contains only $5.1\times 10^3$ particles, which is more than 100 times smaller. 

These experiment results confirm that the proposed gradient domain Gaussian splatting indeed can improve the accuracy of the radiance field and also obtain higher sparsity. 
\subsection{PSNR and Sparsity}
The PSNR improvements on these data sets are summarized in Table~\ref{table}. In general, the gradient domain can improve the accuracy of the Gaussian particle representation.

The improvement might depend on several things, such as the input image number (view number) , image resolution, the complexity of the 3D scene, the light condition, etc. The improvement in the banana data set is high because the input images have a high resolution and the scene has simple geometry. In contrast, the edge guidance in the train and truck data set contains the trees and buildings, which might hamper their Laplacian fields. 
\begin{table} 
	\centering \caption{The PSNR comparison of GDGS and 3DGS. }
	\begin{tabular}{c|c|c|c}\hline
		\hline
		&Banana&Train&Truck\\
		(resolution)&$3008\times2000$&$980\times545$&$979\times546$\\
		(views)&16&301&251\\
		\hline
		3DGS &41.7 &28.0 & 28.4\\
		GDGS &\textbf{ 42.8} &\textbf{28.6 } & \textbf{29.3}\\
		\hline
		\rowcolor{gray!10} improved &1.1  &0.6 & 0.9\\
		\hline
	\end{tabular}
	\label{table}
\end{table}

\begin{table} 
	\centering \caption{The number of particles for GDGS and 3DGS. }
	\begin{tabular}{c|c|c|c}\hline
		\hline
		&Banana&Train&Truck\\
		(resolution)&$3008\times2000$&$980\times545$&$979\times546$\\
		(views)&16&301&251\\
		\hline
		3DGS &$357\times 10^3$ &$372\times 10^3$ & $558\times 10^3$\\
		GDGS &\textbf{ $3\times 10^3$} &\textbf{$2.9\times 10^3$ } & \textbf{$5.1\times 10^3$}\\
		\hline
		\rowcolor{gray!10} speedup &119  &128 & 109\\
		\hline
	\end{tabular}
	\label{table2}
\end{table}

\section{Conclusion}
In this paper, we present a simple yet effective gradient domain Gaussian splatting method. The proposed method works in the gradient and Laplacian domain, instead of the signal domain. Thus, it is sparse and accurate. As shown in the experiments, this method can improve the accuracy of the radian field about $0.6\sim 1$ dB, leading to a much clearer rendering results, especially at edges. Such improvement is important for the view synthesis because the edges contain more visual information~\cite{gong2009symmetry,Gong2023d,Gong2023e}. 

The proposed Gradient Domain Gaussian Splatting method can achieve higher accuracy for the scene representation and rendering. It can be applied in a large range of applications where edge information is important ~\cite{Gong2012,Yu2019,gong2013a,Yin2019a,gong:phd,Yu2022a,gong:cf,Zong2021,Gong2018,Gong2018a,GONG2019329,Yin2019b,Gong2019a,Gong2019,Gong2019c,Gong2022,Yin2020,Gong2020a,Tang2023a,Gong2021a,Gong2021,gong2024eggs}.
\begin{acks}
	This work was supported by National Natural Science Foundation of China (61907031) and Shenzhen Science and Technology Program (20231121165649002 and JCYJ20220818100005011)
\end{acks}
\bibliographystyle{ACM-Reference-Format}
\bibliography{../../IP}

\end{document}